\documentclass[conference]{IEEEtran}
\IEEEoverridecommandlockouts
\usepackage{cite}
\usepackage{amsmath,amssymb,amsfonts}
\usepackage{algorithmic}
\usepackage{graphicx}
\usepackage{textcomp}
\usepackage{xcolor}
\usepackage{booktabs}
\usepackage{multirow} 

\def\BibTeX{{\rm B\kern-.05em{\sc i\kern-.025em b}\kern-.08em
    T\kern-.1667em\lower.7ex\hbox{E}\kern-.125emX}}
\begin{document}

\title{Monocular Endoscopic Tissue 3D Reconstruction with Multi-Level Geometry Regularization
}

\author{
    \IEEEauthorblockN{Yangsen Chen}
    \textit{HKUST(GZ)} \\
    ychen950@connect.hkust-gz.edu.cn
\and
    \IEEEauthorblockN{Hao Wang\textsuperscript{*}}
    \textit{HKUST(GZ)} \\
    haowang@hkust-gz.edu.cn
    \thanks{*Corresponding author.} 
}
\maketitle
\vspace{-10pt}

\begin{abstract}
% This document is a model and instructions for \LaTeX.
% This and the IEEEtran.cls file define the components of your paper [title, text, heads, etc.]. *CRITICAL: Do Not Use Symbols, Special Characters, Footnotes, 
% or Math in Paper Title or Abstract.
Reconstructing deformable endoscopic tissues is crucial for achieving robot-assisted surgery. However, 3D Gaussian Splatting-based approaches encounter challenges in achieving consistent tissue surface reconstruction, while existing NeRF-based methods lack real-time rendering capabilities. In pursuit of both smooth deformable surfaces and real-time rendering, we introduce a novel approach based on 3D Gaussian Splatting. Specifically, we introduce surface-aware reconstruction, initially employing a Sign Distance Field-based method to construct a mesh, subsequently utilizing this mesh to constrain the Gaussian Splatting reconstruction process. Furthermore, to ensure the generation of physically plausible deformations, we incorporate local rigidity and global non-rigidity restrictions to guide Gaussian deformation, tailored for the highly deformable nature of soft endoscopic tissue. Based on 3D Gaussian Splatting, our proposed method delivers a fast rendering process and smooth surface appearances. Quantitative and qualitative analysis against alternative methodologies shows that our approach achieves solid reconstruction quality in both textures and geometries.
\end{abstract}

\begin{IEEEkeywords}
% component, formatting, style, styling, insert
3D Reconstruction, Gaussian Splatting, Robotic Surgery
\end{IEEEkeywords}

\begin{figure*}
    \centering
    \includegraphics[width=0.95\textwidth]{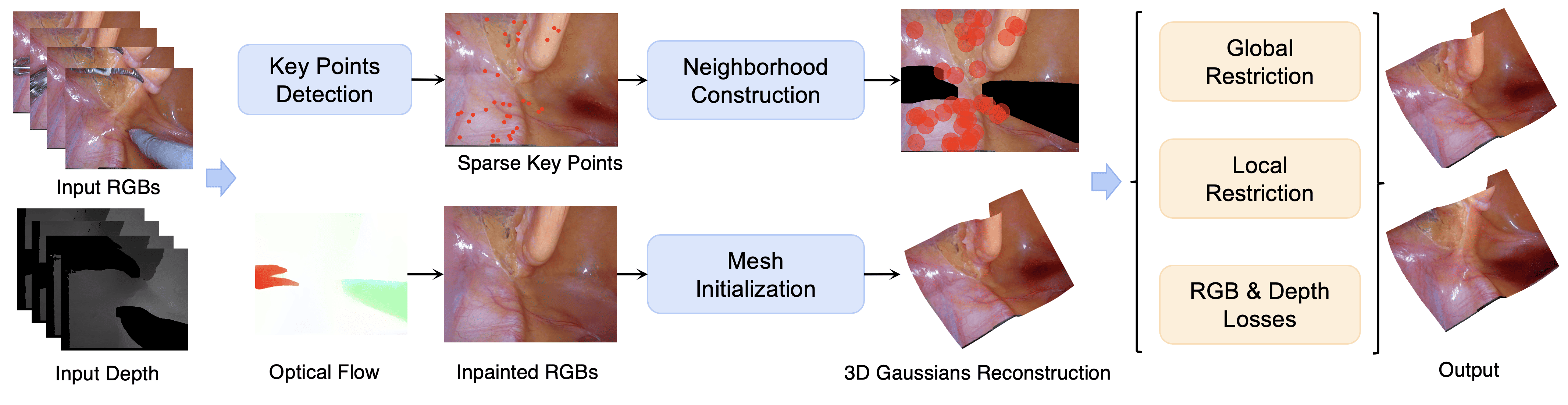}
    % \caption{Our framework. We first do multi modal preparations, then we carry out the surface aware reconstruction, after that the semi-rigidity deformation is performed to get the final result. }
    \caption{Our methodology begins with the key point detection, followed by neighborhood identification and mesh based Gaussian Splatting reconstruction. Subsequently, global and local restriction on the deformation is executed to obtain the final dynamic output.}
    \label{fig:framework} 
\end{figure*}

\section{Introduction}
The 3D reconstruction of surgical scenes from endoscope videos is a critical yet challenging task. It serves as a foundational element in the field of Robot-Assisted Surgery \cite{8918274}, enabling various essential clinical applications~\cite{Han2021ASR, Zhu2022TheRO, li2024endora}.
% These applications include robotic surgery automation, preoperative planning, intra-operative navigation, simulation of surgical environments, immersive medical training, etc.
Despite significant progress in 3D reconstruction techniques for natural scenes \cite{nerf, gs}, endoscopic surgical field reconstruction from videos presents several unresolved challenges.

Firstly, surgical scenarios are typified by deformable structures undergoing significant topological transformations, necessitating the employment of dynamic reconstruction methodologies to accurately encapsulate non-rigidity. While recent advancements in 3D reconstruction techniques \cite{endogs, endonerf, li2024endosparse, liu2024lgs} have exhibited substantial strides in the domain of deformable endoscopic tissues, they exhibit a deficiency in surface-aware constraint, resulting in artifacts. To address this issue, we propose two methodologies: surface-aware reconstruction and semi-rigidity deformation. Our surface-aware reconstruction technique leverages mesh surface constraints to confine Gaussian distributions, thereby yielding seamless outcomes. Furthermore, our semi-rigidity deformation approach is developed based on the pretrained key point detection models. Based on the detected key points from the given images, we integrate local rigidity cues and global non-rigidity cues to steer Gaussian Splatting, enhancing reconstruction fidelity.

Secondly, the limited perspectives captured in endoscopic videos, constrained by camera movements, restrict the availability of 3D cues for reconstructing soft tissue. Furthermore, surgical instruments invariably obstruct portions of the soft tissue, impeding the comprehensive reconstruction of the surgical scene.
To address this challenge, we leverage the existing 2D optical flow detection model. Technically, we produce optical flows from the endoscopic images, which are used to augment and guide the learning process for 3D reconstruction and deformation. To predict occluded regions based on spatial-temporal cues, we introduce a video inpainting module into our framework.
It is notable that our pipeline initially estimates the optical flow of the videos and subsequently incorporates a video inpainting model. We conducted the video inpainting model fine-tuning with a surgical dataset \cite{allan2021stereo}. To our best knowledge, we propose the first work in addressing the spatial-temporal masking issue within this endoscopic tissue 3D reconstruction context.

Moreover, despite the advancements in existing NeRF \cite{nerf} based methods, which have achieved high-quality reconstruction, their limitations include long training time and slow rendering speeds. 
In contrast, our proposed novel Gaussian Splatting-based approach enables real-time rendering and significantly reduces training time. Compared to the concurrent 3D Gaussian-based methodologies, our technique yields solid reconstruction quality in both textures and geometries.

Our contributions in this work are threefold:
\begin{itemize}
    
    \item We propose the surface-aware endoscopic reconstruction, which integrates RGB, depth, and optical flow data, to achieve consistent and smooth geometry reconstructed results.
    
    \item We propose semi-rigidity deformation guidance to model realistic Gaussian deformations through global and local motion learning, which avoids the 3D floaters during the 3D reconstruction process.

    \item We propose a novel approach with multi-level regularization for 3D dynamic endoscopic tissue reconstruction, which demonstrates superior performance in both textures and geometries.

\end{itemize}

\section{Related Works}

\subsection{Endoscopic 3D Reconstruction}

% 对于具有双目相机的系统
Depth estimation based methods such as\cite{hapnet, huoling} explored the effectiveness of surgical scene reconstruction via depth estimation. Since most of the endoscopes are equipped with stereo cameras, depth can be estimated from binocular vision. However these methods can not provide good deformable results.

SLAM-based methods \cite{wang2024endogslam, song2017dynamic,zhou2019real,zhou2021emdq, stilz2024flex} fuse depth maps in 3D space to reconstruct surgical scenes under more complex settings. Nevertheless, these methods either hypothesize scenes as static or surgical tools not present, limiting their practical use in real scenarios. 

Sparse warp field-based methods such as SuPer \cite{super} and E-DSSR \cite{edssr} present frameworks consisting of tool masking, stereo depth estimation to perform singleview 3D reconstruction of deformable tissues. All these methods track deformation based on a sparse warp field \cite{warp}, which is not robust when deformations are significantly beyond the scope of non-topological changes.

With the development of Neural Radiance Field (NeRF) \cite{nerf}, learning based 3D reconstruction has been much more popular, recent works \cite{endonerf, endosurf, lerplane, forplane, based} utilize NeRF for the reconstruction of endoscopic videos. However, due to the implicit representation nature of NeRF, the rendering speed is far from real time, hindering their real world applications.

Some simultaneous works~\cite{endogs, endogaussian, li2024endosparse, liu2024lgs} also used Gaussian Splatting \cite{gs}, while the surface reconstruction accuracy in endoscopic scenes remains challenging for real-world applications.

\section{Methods}

Given a stereo video of deforming tissues, we aim to reconstruct the surface shape $\mathcal{S}$ and texture $\mathcal{C}$.
Similar to EndoNeRF \cite{endonerf}, we take a sequence of frame data $\left\{\left(\mathbf{I}_i, \mathbf{D}_i, \mathbf{M}_i, \mathbf{P}_i \right)\right\}_{i=1}^T$ as input. 
Here $T$ stands for the total number of frames.
$\mathbf{I}_i \in \mathbb{R}^{H \times W \times 3}$ and $\mathbf{D}_i \in \mathbb{R}^{H \times W}$ refer to the $i$-th left RGB image and depth map with height $H$ and width $W$.
Foreground mask $\mathbf{M}_i \in \mathbb{R}^{H \times W}$ is utilized to exclude unwanted pixels, such as surgical tools, blood, and smoke.
Projection matrix $\mathbf{P}_i \in \mathbb{R}^{4 \times 4}$ maps $3 \mathrm{D}$ coordinates to $2 \mathrm{D}$ pixels.
In this work we prioritize 3D reconstruction and deformation.

\subsection{Preparatory Procedures}

In this phase, we initiate preparatory procedures for the training of Gaussian Splatting. We conduct sparse key point matching, intended for subsequent utilization in the semi-rigidity deformation stage, and perform video inpainting to mitigate occlusions caused by surgical instruments, aimed for subsequent application in the surface-aware reconstruction phase.
~\\

\noindent {\bf{Sparse Feature Point Matching.}} We initially identify specific feature points for sparse point tracking. These points are predominantly situated at crucial vascular intersections and regions characterized by distinctive features, making them challenging to reconstruct. We employ the Scale-Invariant Feature Transform (SIFT) \cite{lowe1999object} technique to extract sparse feature points from each frame. Subsequently, we conduct feature matching to ascertain the correspondence of points across frames, thereby establishing the trajectories of sparse key points. The acquisition of sparse tracks furnishes valuable information to facilitate the modeling of tissue deformation. By leveraging the sparse tracks, we can effectively guide the learning process for the deformation dynamics.
~\\

\noindent{\bf{Video Inpainting.}} In this stage, video inpainting is conducted to eliminate occlusions caused by surgical tools. Given the original video sequence of masked surgical tools $X:=\left\{X_1, \ldots, X_T\right\}$, with corresponding annotations of corrupted regions represented by the mask sequence $M:=\left\{M_1, \ldots, M_T\right\}$ (where $T$ denotes the length of the video), our objective is to generate the inpainted video sequence $\hat{Y}:=\left\{\hat{Y}_1, \ldots, \hat{Y}_T\right\}$ while preserving spatio-temporal coherence with the ground truth video sequence $Y:=\left\{Y_1, \ldots, Y_T\right\}$.

Prior approaches have often overlooked the significance of inpainting, resulting in unnatural visual artifacts within the inpainted regions. To address this, we adopt a Transformer\cite{vaswani2017attention}-based inpainting network \cite{fgt} for the video inpainting process. Specifically, we fine-tune a flow-guided video inpainting model to accommodate tool masks, leveraging data from StereoMIS \cite{allan2021stereo}. In this dataset, continuous areas are randomly masked to simulate the occlusion effects of surgical tools. As illustrated in Figure \ref{fig:cut}, our inpainting outcomes exhibit improved visual fidelity.

\subsection{Surface-Aware Reconstruction}

In this phase, our aim is to reconstruct the initial frame of the scene with high quality while being surface-aware. Despite the ability of 3DGS to produce realistic real-time rendered images, it faces challenges in accurately representing the surface of the scene. This challenge arises from the use of discrete Gaussian kernels. However, ensuring precision in depicting the underlying surface of the surgical scene is crucial. To tackle this issue, we incorporate mesh with Gaussian Splatting during the reconstruction of the first frame. Our approach focuses on bounding 3D Gaussian kernels onto the mesh surface, thus facilitating subsequent Gaussian deformation processes. The conceptual foundation for our surface-aware reconstruction is inspired by EndoSuRF (Endoscopic Surgical Reconstruction Framework) \cite{endosurf} and Mesh-based Gaussian Splatting \cite{gao2024meshbased}. Our primary objective in this phase is to achieve a high-quality reconstruction of the first frame.
~\\

\noindent{\bf{Mesh Reconstruction.}}
The initial step involves generating the mesh for the first frame, employing static NeuS2 \cite{neus2} for mesh reconstruction. Each 3D position $\mathbf{x}$ is mapped to its multi-resolution hash encodings $h_{\Omega}(\mathbf{x})$, utilizing learnable hash table entries $\Omega$. Since $h_{\Omega}(\mathbf{x})$ serves as an informative encoding of spatial position, the MLPs responsible for mapping $\mathbf{x}$ to its Signed Distance Function (SDF) $d$ and color $c$ can be kept shallow, ensuring efficient training without compromising quality. The SDF network, denoted as
$
(d, \mathbf{g})=f_{\Theta}(\mathbf{e}),
$ consists of a shallow MLP with weights $\Theta$, where $\mathbf{e}=\left(\mathbf{x}, h_{\Omega}(\mathbf{x})\right)$. Here, $\mathbf{e}$ encapsulates the 3D position $\mathbf{x}$ along with its corresponding hash encoding $h_{\Omega}(\mathbf{x})$, yielding the SDF value $d$ and a geometry feature vector $\mathbf{g} \in \mathbb{R}^{15}$.
The normal vector $\mathbf{n}$ at $\mathbf{x}$ is computed as
$
\mathbf{n}=\nabla_{\mathbf{x}} d,
$
where $\nabla_{\mathbf{x}} d$ represents the gradient of the SDF with respect to $\mathbf{x}$. This normal, combined with the geometry feature $\mathbf{g}$, the SDF $d$, the point $\mathbf{x}$, and the ray direction $\mathbf{v}$, serves as input to the color network, expressed as
$\mathbf{c}=c_{\Upsilon}(\mathbf{x}, \mathbf{n}, \mathbf{v}, d, \mathbf{g}),$
which predicts the color $\mathbf{c}$ of $\mathbf{x}$.

% 我们使用了color和depth+eikonal的监督
To supervise the learning of NeuS2, we minimize the color difference between the rendered pixels $\hat{C}_i$ with $i \in\{1, \ldots, m\}$ and the corresponding ground truth pixels $C_i$ and also minimize the depth difference between ground truth depth and predicted depth: 
\begin{equation}
\mathcal{L}_{\text {color }}=\frac{1}{m} \sum_i \mathcal{R}\left(\hat{C}_i, C_i\right)
, 
\mathcal{L}_{\text {depth }}=\frac{1}{m} \sum_i \mathcal{R}\left(\hat{D}_i, D_i\right)    
\end{equation}

where $\mathcal{R}$ is the Huber loss. We also employ an Eikonal term 
\begin{equation}
    \mathcal{L}_{\text {eikonal }}=\frac{1}{m n} \sum_{k, i}\left(\left\|\mathbf{n}_{k, i}\right\|-1\right)^2
\end{equation}
to regularize the learned signed distance field, where $k$ indexes the $k$-th sample along the ray with $k \in\{1, \ldots, n\}$, $n$ is the number of sampled points, and $\mathbf{n}_{k, i}$ is the normal of a sampled point. Our final loss for the mesh reconstruction of the first frame:
\begin{equation}
    \mathcal{L_{\text {mesh }}}=\mathcal{L}_{\text {color }}+\alpha_{1} \mathcal{L}_{\text {depth }}+\alpha_{2} \mathcal{L}_{\text {eikonal}}
\end{equation}
~\\
\noindent{\bf{Mesh restricted Gaussian Splatting.}}
Upon acquiring the mesh, Gaussian kernels are positioned at the centroid of each mesh triangle, establishing a direct correspondence between Gaussian kernels and mesh triangles. The initial radius aligns with the size of the inscribed circle within the binding triangle. The initial Gaussian training proceeds for the first frame without heuristics such as deletion or splitting of Gaussians.

To enhance the visual fidelity of 3D Gaussians, a regularization process is implemented to maintain spatial coherence and local consistency. This regularization mitigates potential visual distortions arising from overly expansive Gaussians that cover multiple mesh triangles. To ensure the fidelity of deformation outcomes, a regularization term \( L_{\text{scale}} \) is introduced, the formulation of this regularization term is expressed as:

\begin{equation}
L_{\text{scale}}=\frac{1}{ |\mathcal{G}|}  \sum_{g_{i} \in \mathcal{G}} \max \left(\max \left(s_i\right)-\gamma_1 R_i, 0\right)
\end{equation}

where $s_i$ represents the 3D scaling vector of each Gaussian, $R_i$ denotes the radius of the circumcircle of the binding triangle wherein the Gaussian is positioned, and $\gamma_1$ denotes the hyperparameter. This loss dynamically adjusts the Gaussian size relative to the binding triangle's radius during training. This method ensures the acquisition of suitable Gaussian representations and maintains local continuity during deformation. 

Additionally, we impose constraints on the displacement of Gaussians, preventing them from shifting away from the binding triangle:
\begin{equation}
L_{\text{shift}}= \frac{1}{ |\mathcal{G}|}  \sum_{g_{i} \in \mathcal{G}} \max \left(\max \left(\Delta\mu \right)-\gamma_2 R_i, 0\right)
\end{equation}
where $\Delta \mu$ denotes the shifted distance, and $\gamma_2$ represents the hyper-parameters. This loss penalizes large shifts for Gaussians. Therefore, the comprehensive loss function for the first frame Gaussian Splatting is formulated as

\begin{equation}
    \mathcal{L}_{\text {reconstruction}} =\mathcal{L}_{\text {color }}+\beta_{1} \mathcal{L}_{\text {depth }}+\beta_{2} \mathcal{L}_{\text {scale}} +\beta_{3} \mathcal{L}_{\text {shift}}
\end{equation}

\subsection{Semi-Rigidity Deformation}
\label{sec:tracking}

After achieving high-quality reconstruction of the initial frame, we advocate for semi-rigidity deformation in the training of subsequent frames using Gaussian Splatting to preserve multi-level geometry regularization. This module is designed to facilitate the acquisition of physically plausible deformations. To address this objective, we introduce two guiding methodologies: local rigidity restriction and global non-rigidity restriction. Our local rigidity restriction aims to guide the learning on the area where there exists key point features, and our global non-rigidity restriction aims to unify the global deformation.
~\\

\begin{figure*}[t]
    \centering
    \includegraphics[width=.95\textwidth]{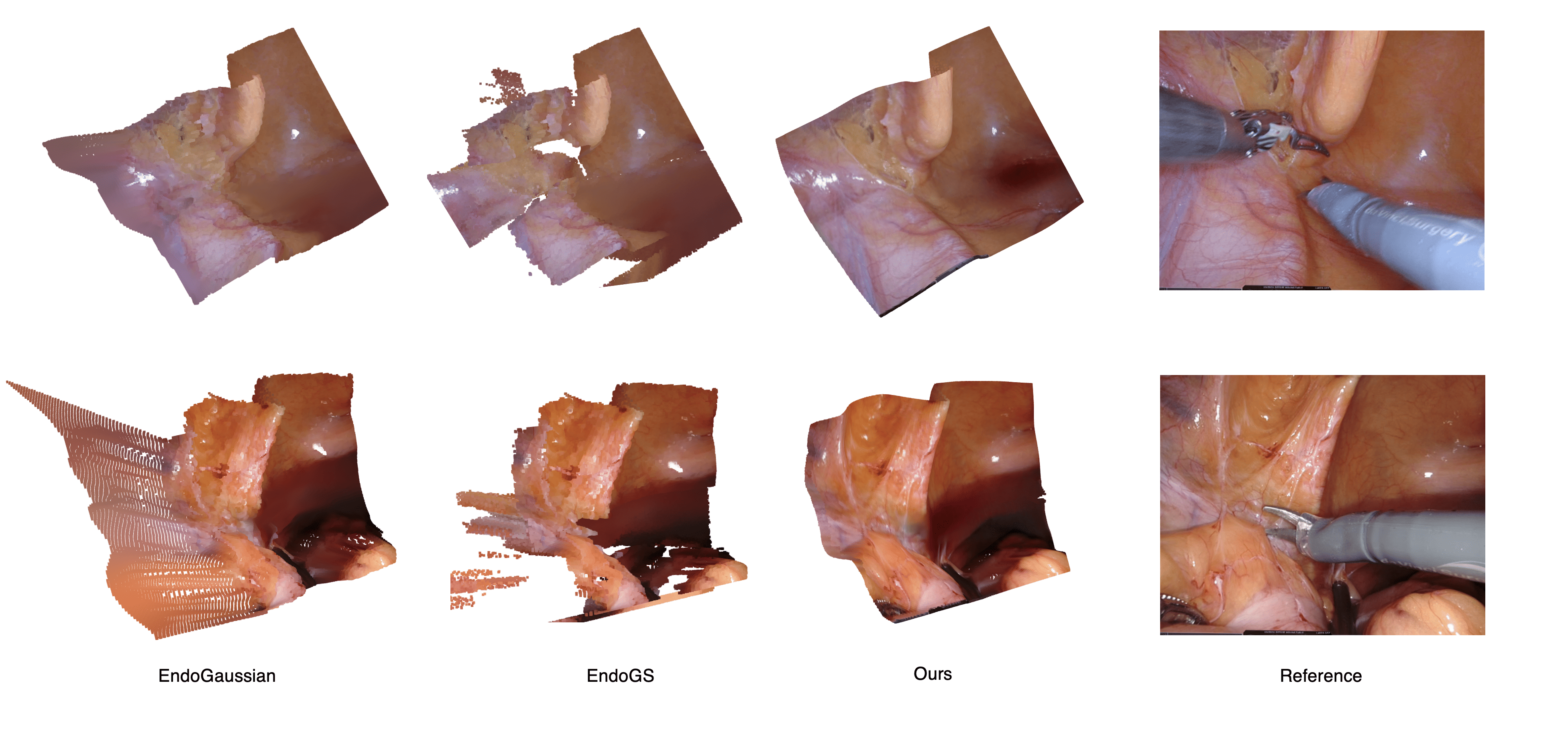}
    \caption{
Comparison with other Gaussian Splatting-based methodologies. When we change the viewpoints, we clearly observe the 3D geometries of existing works are distorted, while our proposed framework reconstructs more consistent and much smoother endoscopic tissue surfaces. This demonstrates the usefulness of our proposed multi-level geometry regularization. 
}
    \label{fig:threed} 
    \vspace{-10pt}

\end{figure*}

\noindent {\bf{Local Rigidity Restriction.}} 
We employ a methodology inspired by the as-rigid-as-possible (ARAP) approach \cite{arap} for mesh deformation. Since the strict adherence to rigidity may lead to inaccuracies in the resulting depth, we introduce an ARAP Loss $\mathcal{L}_{\text{arap}}$ to gently guide points within local regions to conform to the principle of maximal rigidity. Given that endoscopic tissues exhibit high deformability, rigidity is primarily localized to small areas. As depicted in Figure \ref{fig:framework}, we define the region as within a specified distance from the key points; only within these designated areas is the ARAP loss implemented.

To compute $\mathcal{L}_{\text{arap}}$, we utilize the points at the current time step $t$ and the corresponding points at the previous time step $t-1$. For each point $i$ within the region influenced by the ARAP Loss, its nearest sparse key point is represented as $k$. Initially, we compute the rotation matrix:
\begin{equation}
\hat{R}_i=\underset{R \in \mathbf{S O}(3)}{\arg \min } \sum_{k \in \mathcal{K}} w_{i k}\left\|\left(\mu_i^{t}-\mu_k^{t}\right)-R\left(\mu_i^{t-1}-\mu_k^{t-1}\right)\right\|^2
\end{equation}
This rotation matrix can be calculated easily by using SVD decomposition. After getting the rotation matrix, we can define our ARAP Loss ($\mathcal{L}_{\text {arap}}$) as:
\begin{equation}
\frac{1}{n |\mathcal{S}|}  \sum_{i \in \mathcal{S}}\sum_{k \in \mathcal{K}} w_{ik}\left\|\left(\mu_{i, t}-\mu_{k, t}\right)-\hat{R}_i\left(\mu_{i, t-1}-\mu_{k, t-1}\right)\right\|^2
\end{equation}
% 这评估了学习运动偏离局部刚性假设的程度。
where n denotes the number of key points, $\mathcal{S}$ denotes all the points restricted by the loss, $\mathcal{K}$ is the group of key points, $w_{ik}$ is the contagent as described in \cite{arap}. This loss evaluates the degree to which the learned motion deviates from the assumption of local rigidity principle described by the ARAP. By penalizing $\mathcal{L}{\text {arap}}$, the learned motions are encouraged to be locally rigid.
~\\

\noindent{\bf{Global Non-Rigidity Restriction.}}
Since the key points are limited and the radius of our ARAP regulation cannot cover all points, we use neighborhood similarity loss to handle changes globally. We explicitly encourage the neighbouring Gaussians to have the same rotation over time:
\begin{equation}
    \mathcal{L}_{\text{rot}} = \frac{1}{r |\mathcal{S}|} \sum_{i \in \mathcal{S}} \sum_{ j \in \text{knn}_{i;r}} \left\| \hat{q}^{}_{j,t} \hat{q}^{-1}_{j,t-1} - \hat{q}^{}_{i,t} \hat{q}^{-1}_{i,t-1}  \right\|_2
\end{equation}

where $\hat{q}$ denotes the normalized quaternion representation of each Gaussian's rotation, with \( j \) belonging to the neighborhood of \( i \) as determined by the k-nearest-neighbor criterion. Unlike other methodologies such as that proposed by \cite{scgs}, which necessitates additional computational steps for neighborhood determination, our approach obviates the need for further classifications of nearby neighbors. This is attributed to the clarity in neighborhood identification facilitated by our surface-aware reconstruction..

 We apply $\mathcal{L}_{\text{rot}}$ only between the current timestep and the directly preceding timestep, thus only enforcing these losses over short-time horizons, this sometimes causes elements of the scene to drift apart, thus we apply another loss, the isometry loss, over the long-term:
\begin{equation}
    \mathcal{L}_{\text{iso}}\hspace{-0.5ex}=\hspace{-0.5ex}\frac{1}{r |\mathcal{S}|} \sum_{ i \in \mathcal{S}} \sum_{ j \in \text{knn}_{i;r}}\hspace{-1.5ex}w_{i,j} \left| \left\| \mu_{j,0} - \mu_{i,0} \right\|_2\hspace{-0.5ex}-\left\| \mu_{j,t} - \mu_{i,t} \right\|_2 \right|
\end{equation}

This is a weaker constraint in that instead of enforcing the positions between two Gaussians to be the same it only enforces the distances between them to be the same. 

Therefore, combining global and local constraint, our overall loss for deformable Gaussian Splatting is defined as:
\begin{equation}
    \mathcal{L}_{\text{deform}}= \mathcal{L}_{\text{color}} + \lambda_{1}\mathcal{L}_{\text{depth}} + \lambda_{2}\mathcal{L}_{\text{arap}} + \lambda_{3}\mathcal{L}_{\text{rot}} + \lambda_{4}\mathcal{L}_{\text{iso}}
\end{equation}

\begin{figure*}[ht!]
    \centering
    \includegraphics[width=.85\textwidth]{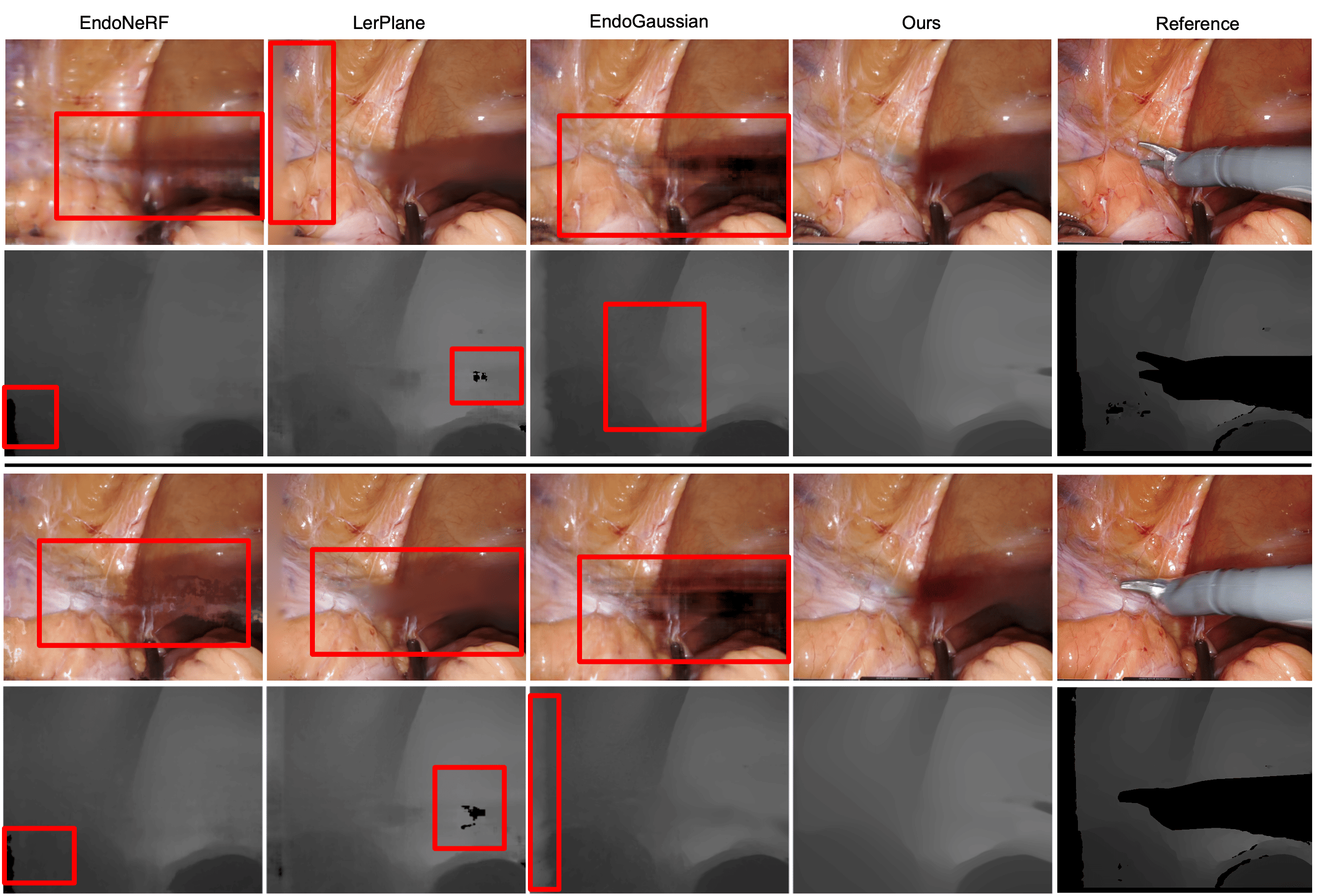}
    \caption{Comparison with other methods on image quality and depth quality, where red boxes denote the incorrect reconstructed areas. Our method presents better reconstructed textures in the occluded regions.
    }
    
    \label{fig:cut} 
\end{figure*}

\section{Experiments}

\begin{table*}[t]

\caption{Comparison on the ENDONERF dataset \cite{endonerf}. Our approach demonstrates superior performance across most metrics and scenarios.}

\begin{center}
\begin{tabular}{l|ccc|ccc|ccc}
\toprule \multirow{2}{*}{ Models } & \multicolumn{3}{c|}{Cutting} & \multicolumn{3}{c|}{Pulling}  &  \multirow{2}{*}{ Training Time $\downarrow$}& \multirow{2}{*}{FPS $\uparrow$}   & \multirow{2}{*}{GPU Usage $\downarrow$}  \\

& PSNR $\uparrow$ & SSIM $\uparrow$& LPIPS $\downarrow$ & PSNR $\uparrow$ &SSIM $\uparrow$ &LPIPS $\downarrow$ &   \\

\midrule 
EndoNeRF &  35.64 & 0.930 & 0.132  & 34.71 & 0.920 & 0.095  & $\sim$ 6 hrs & $\sim$ 0.2 & $\sim$ 20GB \\

EndoSurf & 35.89 & 0.952 & 0.107 & 34.91 &  0.955 &  0.120& $\sim$ 7 hrs& $\sim$ 0.04 &  $\sim$ 20GB\\

LerPlane & 34.69 & 0.901 & 0.112 & 36.38 & 0.937 & 0.083 & $\sim$ 8 min & $\sim$ 0.9 & $\sim$ 20GB \\

EndoGS & 36.20 & 0.958 &\textbf{ 0.044} & 38.21 &\textbf{ 0.967} & 0.066 & \textbf{$\sim$ 2 min} & $\sim$ 60 &  $\sim$ 10GB  \\

EndoGaussian & 37.21 & 0.961 & 0.065 & 36.10& 0.946 & 0.091 & \textbf{$\sim$ 2 min} &\textbf{ $\sim$ 170} & \textbf{ $\sim$ 3GB} \\

\midrule 

Ours & \textbf{38.05 }& \textbf{0.965} & 0.047 & \textbf{38.27} & 0.951 & \textbf{0.046} & \textbf{$\sim$ 2 min}& \textbf{$\sim$ 170 } &  \textbf{$\sim$ 3GB} \\

\bottomrule
\end{tabular}

\end{center}

\label{tab:endonerf}

\end{table*}

\begin{table}[t]

\caption{Comparison on the SCARED dataset \cite{allan2021stereo}.}
\begin{center}
\begin{tabular}{l|ccc}
\toprule 
 Models & PSNR $\uparrow$ & SSIM $\uparrow$& LPIPS $\downarrow$  \\
 \midrule

EndoNeRF \cite{endonerf} &  24.34 & 0.759 & 0.320  \\

EndoSurf \cite{endosurf} & 25.02 & 0.802 & 0.356 \\

EndoGS \cite{endogs} & 26.47 & 0.798 & 0.291 \\

EndoGaussian \cite{endogaussian} & 27.04 & \textbf{0.825} & \textbf{0.275 }\\
\midrule 

Ours &\textbf{ 28.31 }& 0.810 & 0.282  \\

\bottomrule
\end{tabular}

\end{center}

\label{tab:scared}
\end{table}

\begin{figure*}[t]
    \centering
    \includegraphics[width=.80\textwidth]{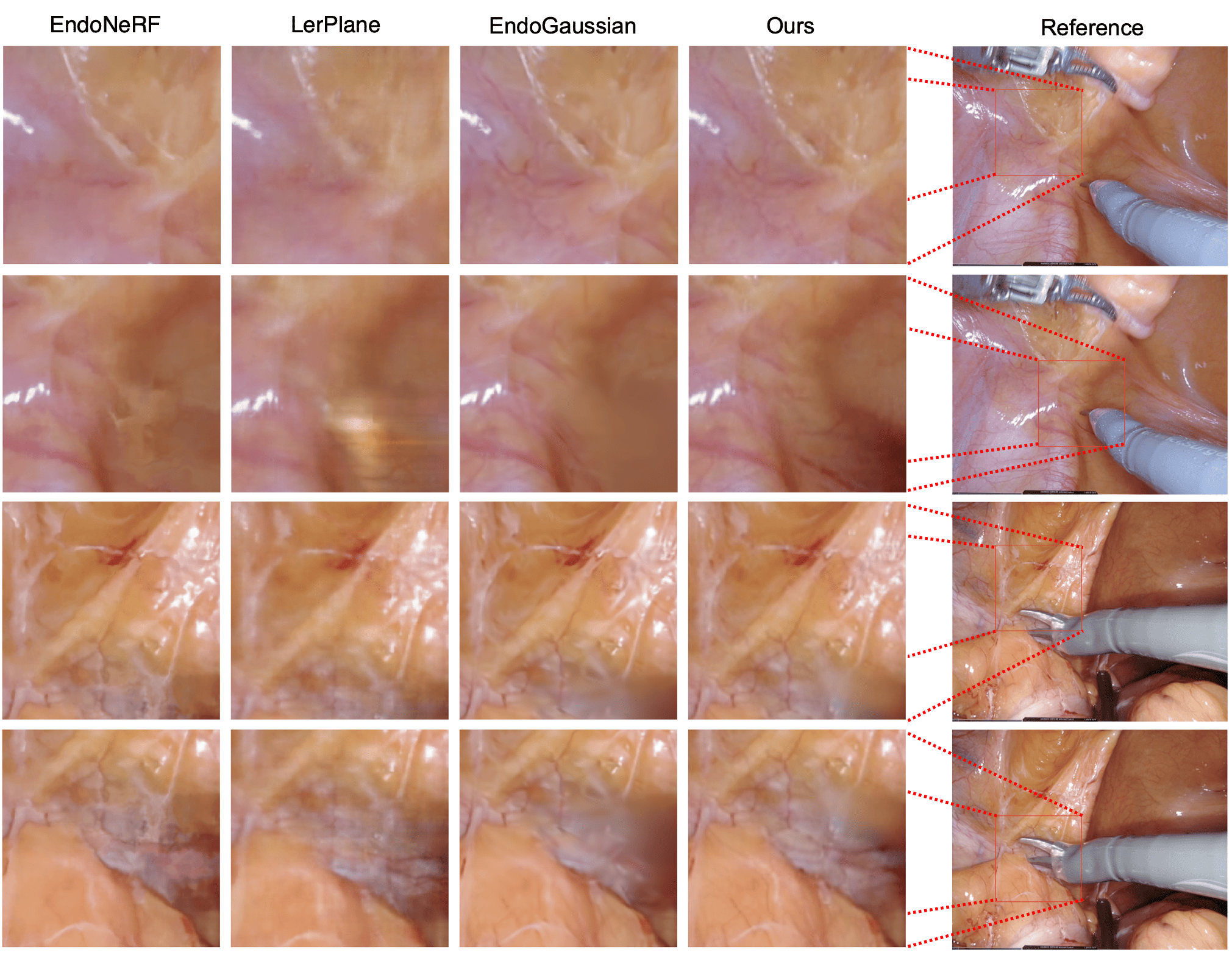}
    
    \caption{Detailed comparisons of the output RGB image are presented in this section. We emphasize regions rich in blood vessels and intricate texture. Our approach demonstrates a notable capability in preserving intricate details when contrasted with alternative methodologies. Furthermore, the inpainted regions in our method exhibit a higher degree of realism and coherence.}
    \label{fig:inpaint} 
\end{figure*}

\subsection{Datasets}

We conduct experiments utilizing two publicly available datasets: ENDONERF \cite{endonerf} and SCARED \cite{allan2021stereo}.
The EndoNeRF dataset \cite{endonerf} provides two instances of in-vivo prostatectomy data. This dataset includes depth maps estimated using E-DSSR \cite{edssr}, accompanied by manually labeled tool masks. ENDONERF\cite{endonerf} encompasses two cases of in-vivo prostatectomy data captured from stereo cameras at a single viewpoint. It presents challenging scenes featuring non-rigid deformation and tool occlusion.
The SCARED dataset \cite{allan2021stereo} consists of ground truth RGBD images of five porcine cadaver abdominal anatomies. SCARED gathers RGBD images of the same anatomies using a DaVinci endoscope and a projector.
~\\

\subsection{Main Results}

In our experiments, the loss weight coefficients were determined via a coarse grid search on the validation set. For the mesh reconstruction stage (NeuS2), we set the depth loss weight to $\alpha_1 = 0.1$ and the Eikonal term weight to $\alpha_2 = 0.01$. In the mesh-restricted Gaussian Splatting, we used $\beta_1 = 0.5$ for the depth loss, $\beta_2 = 0.1$ for the scale regularization loss, and $\beta_3 = 0.05$ for the shift regularization loss. For the deformation stage, the loss terms were weighted as follows: $\lambda_1 = 0.5$ for the depth loss, $\lambda_2 = 0.1$ for the local ARAP loss, $\lambda_3 = 0.05$ for the rotation consistency loss, and $\lambda_4 = 0.02$ for the isometry loss. Variations of $\pm20\%$ in these values resulted in only marginal performance changes, indicating the robustness of our settings.

We performed a comparative analysis of our proposed approach with other techniques: EndoNeRF \cite{endonerf}, EndoSurf \cite{endosurf}, LerPlane \cite{lerplane}, EndoGS, and EndoGaussian \cite{endogaussian}. Among these, EndoNeRF, EndoSurf, and LerPlane are methods based on NeRF, while EndoGS and EndoGaussian utilize Gaussian Splatting.

As depicted in Table~\ref{tab:endonerf} and Table~\ref{tab:scared}, while methodologies rooted in NeRF, such as EndoNeRF and EndoSurf, demonstrate the ability to generate reconstructions of deformed tissues at a high-quality level, they come with a significant training cost, requiring hours of optimization and substantial memory resources. LerPlane \cite{lerplane} notably reduces the training duration to approximately 3 minutes per frame but compromises on the fidelity of reconstructions. Although successive iterations of LerPlane have the potential to augment rendering quality, they still encounter impediments in terms of inference speed. Their overall image quality falls short in comparison to Gaussian Splatting-based methodologies.

Comparing with other Gaussian Splatting-based methodologies, as depicted quantitatively in Tables \ref{tab:endonerf} and \ref{tab:scared}, our approach demonstrates superior performance across three key metrics: PSNR (Peak Signal-to-Noise Ratio), SSIM (Structural Similarity Index Measure), and LPIPS (Learned Perceptual Image Patch Similarity). Furthermore, upon qualitative examination, our model exhibits enhanced 3D visual fidelity, as illustrated in Figure \ref{fig:threed}. In contrast, other Gaussian-based techniques exhibit deficiencies such as the presence of artifacts and suboptimal surface reconstruction. Our method achieves smoother surface representations, as evidenced in the visual depiction.

Our approach achieves state-of-the-art reconstruction outcomes (PSNR of 38.162) within a 2 minutes per frame of training, and our method achieves real-time rendering rates exceeding 60 frames per second (FPS), signifying a substantial acceleration compared to NeRF based techniques. Additionally, we note that our method utilizes only 2GB of GPU memory for optimization, approximately one-tenth the consumption of prior techniques, thereby mitigating hardware prerequisites for implementation in surgical settings.
To provide a more intuitive comparison, we present several qualitative results in Figure \ref{fig:cut} and \ref{fig:inpaint}. It is evident from these results that our method preserves more details and offers superior visualization of deformable tissues compared to other methods.

To compare with other methods for the Gaussian Splatting reconstruction of the first frame, we first experimented on the efficacy of other Gaussian Splatting based reconstruction methods : the original 3DGS, Dynamic 3D Gaussians \cite{dg}, Depth-Regularized GS \cite{chung2024depthregularized} and FS-GS \cite{fsgs}, and our proposed surface-aware reconstruction technique. The results are outlined in Table \ref{tab:else}.

When using the original 3DGS method, the result is undesirable, with the PSNR only at 25.15, SSIM at 0.83 and LPIPS at 0.536. This deficiency is mainly due to its reliance on single-view data. We further train the original 3DGS with additional depth guidance, but simple adding depth guidance do not lead to better reconstruction quality.

For FS-GS, Depth-Regularized GS and Dynamic 3D Gaussians, although they demonstrate an improvement in reconstruction quality. However, despite the enhancements, the outcome remained not competitive with our surface-aware reconstruction. This is due to none of these methods is designed for single view situation, while in endoscopic scenario, there only exists single view.
In comparison, our surface-aware reconstruction technique provides higher quantitative results in terms of all three metrics.

% To further investigate alternative approaches, we examined the Dynamic 3D Gaussians method proposed by \cite{dg}. This method, like ours, aims to learn deformation through Gaussian splatting. However, when we substituted our deformation learning module with this method and evaluated its performance in the context of endoscopic single view scenes, the reconstruction quality is limited. As shown in Table \ref{tab:ab}, the Dynamic 3D Gaussians method struggled to adapt to endoscopic specific requirements compared to our approach.

\begin{table}[tbp]
% \scriptsize
\centering
\caption{Comparison on first frame 3D reconstruction quality.}

\begin{tabular}{l|ccc}
\toprule
 Method & PSNR$\uparrow$ & SSIM$\uparrow$ & LPIPS$\downarrow$  \\
\midrule

3DGS \cite{gs} & 25.15 & 0.803 & 0.536 \\
Dynamic 3D Gaussians \cite{dg} & 33.60 & 0.914 & 0.096  \\
Depth-Regularized GS \cite{chung2024depthregularized} & 28.38 & 0.913  & 0.472 \\
FS-GS \cite{fsgs} & 30.24 & 0.927 & 0.124 \\
\midrule
Ours & \textbf{38.16 }& \textbf{0.965} & \textbf{0.046}  \\

\bottomrule
\end{tabular}

\label{tab:else}
\end{table}

\subsection{Ablation Study}

We also carried out the ablation study on our proposed components, the Surface-Aware Reconstruction and the Semi-Rigidity Deformation.

% For the Surface-Aware Reconstruction method, when we do not use this method, our framework faces a degrade on overall image quality, with PSNR at 33.60, SSIM at 0.914 and LPIPS at 0.096.
% For the Semi-Rigidity Deformation method, we conducted an ablation study focusing on two key components: the local rigidity restriction and the global non-rigidity restriction. This analysis revealed that these components are crucial for maintaining image quality. Without them, our model experiences degradation in image quality.
% When the local rigidity restriction is omitted, there exists degradation in the overall image quality across the three metrics. Similarly, neglecting the global non-rigidity restriction also results in a degradation of image quality, attributed to the irregular motion of Gaussians according to our examinations.
For the Surface-Aware Reconstruction, this component plays a crucial role in ensuring accurate and smooth surface representation by constraining the 3D Gaussian kernels onto the mesh. When we remove this module, our framework exhibits a significant degradation in overall image quality. Specifically, as presented in Table \ref{tab:ab}, the PSNR drops to 33.60, SSIM decreases to 0.914, and LPIPS increases to 0.096. These results highlight that without the surface-aware reconstruction, our model struggles to effectively represent the underlying geometry, leading to less accurate surface reconstructions with visible artifacts. This is particularly evident in highly deformable regions, where the lack of mesh constraints causes the Gaussians to drift and fail to form a coherent surface.

For the Semi-Rigidity Deformation, we performed a detailed ablation study on its two key components: the local rigidity restriction and the global non-rigidity restriction. Both components are crucial for preserving the structural consistency of the reconstructed scene during deformation. When the local rigidity restriction is omitted, the overall image quality degrades, with PSNR dropping to 37.18, SSIM to 0.950, and LPIPS increasing to 0.065. This degradation is primarily due to the lack of guidance in regions with key points, which results in less accurate motion learning and leads to inconsistencies in deformable regions, especially around areas with vascular intersections or fine features. 

Similarly, when the global non-rigidity restriction is removed, the model suffers from irregular motion of the Gaussians, causing further degradation in the reconstruction quality. In this case, the PSNR drops to 36.29, SSIM to 0.944, and LPIPS increases to 0.063. The absence of this global constraint leads to a lack of cohesion in the deformation process, causing neighboring Gaussians to move inconsistently and resulting in artifacts or distortions in the reconstructed surfaces.

\begin{table}[tbp]

\caption{Ablation study  of our proposed multi-level geometry regularization on the ENDONERF dataset \cite{endonerf}.}

\centering

\begin{tabular}{l|ccc}
\toprule
Method & PSNR$\uparrow$ & SSIM$\uparrow$ & LPIPS$\downarrow$  \\
\midrule
Ours & \textbf{38.16} & \textbf{0.965} & \textbf{0.046} \\
\midrule
w/o surface-aware reconstruction & 33.60 & 0.914 & 0.096  \\
w/o global non-rigidity restriction & 36.29 & 0.944 & 0.063 \\
w/o local rigidity restriction & 37.18 & 0.950 & 0.065  \\

\bottomrule
\end{tabular}

\label{tab:ab}

\end{table}

\section{Conclusion}

In this paper, we propose a novel 3D Gaussian Splatting \cite{gs} based framework with multi-level geometry regularization for real-time and high-quality reconstruction of dynamic endoscopic scenes. By employing surface-aware reconstruction and semi-rigidity deformation, we address the challenge of reconstructing deformable tissue. Experimental results have demonstrated that our method achieves state-of-the-art reconstruction quality, with smooth surfaces and realistic deformation. Additionally, we achieve real-time rendering speeds over 100 times faster than previous NeRF \cite{nerf} based methods, with training times reduced by a factor of 10. We believe that our approach utilizing Gaussian Splatting-based reconstruction techniques can inspire advancements in robotic surgery scene reconstruction.

\section{Acknowledgement}

This research is supported by the National Natural Science Foundation of China (No. 62406267), Guangzhou-HKUST(GZ) Joint Funding Program (Grant No.2025A03J3956), Education Bureau of Guangzhou Municipality and the Guangzhou Municipal Education Project (No. 2024312122).

\bibliographystyle{IEEEtran}
\bibliography{sample-base}

\end{document}